\documentclass[acmtog,nonacm]{templates/acmart/acmart}

\newcommand{\status}{1}

\input{conf/packages}
\definecolor{INCOMPLETECOLOR}{RGB}{178,34,34}
\definecolor{UNDERREVISIONCOLOR}{RGB}{210,121,121}
\definecolor{FEEDBACKNEEDEDCOLOR}{RGB}{230,170,50}
\definecolor{FEEDBACKGIVENCOLOR}{RGB}{121,210,121}
\definecolor{COMPLETECOLOR}{RGB}{121,124,210}
\definecolor{LOCKEDCOLOR}{RGB}{153,102,255}

\definecolor{TODOCOLOR}{RGB}{255,0,0}
\definecolor{MONDECOLOR}{RGB}{0,0,255}
\definecolor{QICOLOR}{RGB}{118,185,0}
\definecolor{ANJULCOLOR}{RGB}{127,127,0}
\definecolor{RACHELCOLOR}{RGB}{127,0,127}
\definecolor{GUESTCOLOR}{RGB}{0,127,127}

\definecolor{WHITE}{RGB}{255,255,255}

\newcommand{\nothing}[1]{}
\newcommand{\isolated}[1]{\hfill\break#1\xspace}

\ifthenelse{\equal{\status}{0}} {                     %

    \newcommand{\todo}[1]{%
        \addcontentsline{toc}{subsection}{%
            \protect\numberline{}%
            \textcolor{TODOCOLOR}{[TODO] #1}}%
            \textcolor{TODOCOLOR}{[TODO] \emph{#1}}}%
    \newcommand{\warning}[1]{\todo{#1}}
    \newcommand{\note}[1]{{\it\color{blue} #1}}

    \newcommand{\todolist}{\newpage\tableofcontents}

    \newcommandx{\monde}[2][1=]
        {\setulcolor{MONDECOLOR}{\ul{#1}}
         \isolated{\textcolor{MONDECOLOR}{\textbf{Monde:} #2}}}
    \newcommandx{\qisun}[2][1=]
        {\setulcolor{QICOLOR}{\ul{#1}}
         \isolated{\textcolor{QICOLOR}{\textbf{Qi:} #2}}}
    \newcommandx{\anjul}[2][1=]
        {\setulcolor{ANJULCOLOR}{\ul{#1}}
         \isolated{\textcolor{ANJULCOLOR}{\textbf{Anjul:} #2}}}
    \newcommandx{\rachel}[2][1=]
        {\setulcolor{RACHELCOLOR}{\ul{#1}}
         \isolated{\textcolor{RACHELCOLOR}{\textbf{Rachel:} #2}}}

    \newcommandx{\guest}[3][1=]
        {\setulcolor{LOCKEDCOLOR}{\ul{#1}} \textcolor{LOCKEDCOLOR}
        {[\textbf{#2:} #3]}}
}{                                                    %
    \newcommand{\todo}[1]{}
    \newcommand{\warning}[1]{}
    \newcommand{\note}[1]{}
    \newcommand{\todolist}{}
    \newcommandx{\monde}[2][1=]{#1}
    \newcommandx{\qisun}[2][1=]{#1}
    \newcommandx{\anjul}[2][1=]{#1}
    \newcommandx{\rachel}[2][1=]{#1}
    \newcommandx{\guest}[3][1=]{#1}
}

\ifthenelse{\equal{\status}{0}} {          %
    
}{}
\ifthenelse{\equal{\status}{1}} {          %
    
}{}
\ifthenelse{\equal{\status}{2}} {          %
    
}{}
\ifthenelse{\equal{\status}{3}} {          %
    
}{}
\ifthenelse{\equal{\status}{4}} {          %
    
}{}

\ifthenelse{\equal{\status}{0}} {
    \newcommand{\badge}[2]{\colorbox{#1}{\small\textcolor{WHITE}{\texttt{#2}}}}
    \newcommand{\headerBadge}[2]{\hspace*{\fill}\badge{#1}{#2}}

    \newcommand{\feedbackNeeded}{\headerBadge{FEEDBACKNEEDEDCOLOR}{feedback needed}}

}{
    \newcommand{\badge}[2]{}{}
    \newcommand{\headerBadge}[2]{}{}

    \newcommand{\feedbackNeeded}{}

}

\newcommand{\veryshortarrow}[1][3pt]{\mathrel{%
   \hbox{\rule[\dimexpr\fontdimen22\textfont2-.2pt\relax]{#1}{.4pt}}%
   \mkern-4mu\hbox{\usefont{U}{lasy}{m}{n}\symbol{41}}}}
\newcommand{\diff}{\, \mathrm{d}} %

\newcommand{\cbm}{\bm{c}}
\newcommand{\ebm}{\bm{e}}
\newcommand{\sbm}{\bm{s}}
\newcommand{\xbm}{\bm{x}}
\newcommand{\gbm}{\bm{g}}

\newcommand{\rbm}{\bm{r}}

\newcommand{\Lbm}{\bm{L}}

\newcommand{\mbm}{\bm{m}}

\newcommand{\thetabm}{\bm{\theta}}
\newcommand{\mubm}{\bm{\mu}}

\newcommand{\Dbm}{\bm{D}}

\newcommand{\Sbm}{\bm{S}}
\newcommand{\Pbm}{\bm{P}}
\newcommand{\Wbm}{\bm{W}}

\newcommand{\Ibm}{\bm{I}}
\newcommand{\Rbm}{\bm{R}}
\newcommand{\Jbm}{\bm{J}}
\newcommand{\Sigmabm}{\bm{\Sigma}}

\newcommand{\R}{\mathbb{R}}

\newcommand{\Ical}{\mathcal{I}}

\newcommand{\Lcal}{\mathcal{L}}
\newcommand{\Dcal}{\mathcal{D}}

\newcommand{\Ecal}{\mathcal{E}}

\newcommand{\Ucal}{\mathcal{U}}

\newcommand{\gs}{3DGS}

\newcommand{\nerf}{NeRF}
\newcommand{\ours}{Ours}

\title{High-Speed Dynamic 3D Imaging with Sensor Fusion Splatting}

\author{Zihao Zou}
\email{zihaozou@cs.unc.edu}
\affiliation{%
  \institution{University of North Carolina, Chapel Hill}
  \streetaddress{232 S Columbia St}
  \city{Chapel Hill}
  \state{NC}
  \country{USA}
  \postcode{27514}
}

\author{Ziyuan Qu}
\email{quinton.qu.gr@dartmouth.edu}
\affiliation{%
  \institution{Dartmouth College}
  \city{Hanover}
  \state{NH}
  \country{USA}
  \postcode{03755}
}

\author{Xi Peng}
\email{xipeng@cs.unc.edu}
\affiliation{%
  \institution{University of North Carolina, Chapel Hill}
  \streetaddress{232 S Columbia St}
  \city{Chapel Hill}
  \state{NC}
  \country{USA}
  \postcode{27514}
}

\author{Vivek Boominathan}
\email{vivekb@rice.edu}
\affiliation{%
 \institution{Rice University}
 \city{Houston}
 \state{TX}
 \country{USA}
 \postcode{77005}
}

\author{Adithya Pediredla}
\email{adithya.k.pediredla@dartmouth.edu}
\affiliation{%
  \institution{Dartmouth College}
  \city{Hanover}
  \state{NH}
  \country{USA}
  \postcode{03755}
}

\author{Praneeth Chakravarthula}
\email{cpk@cs.unc.edu}
\affiliation{%
  \institution{University of North Carolina, Chapel Hill}
  \streetaddress{232 S Columbia St}
  \city{Chapel Hill}
  \state{NC}
  \country{USA}
  \postcode{27514}
}

\renewcommand{\shortauthors}{Zou, Qu, Peng, Boominathan, Pediredla, Chakravarthula}

\begin{document}
\begin{teaserfigure}
  \centering
  \subfloat[Sensor Fusion Splatting Overview]{
    \includegraphics[width=0.42\linewidth]{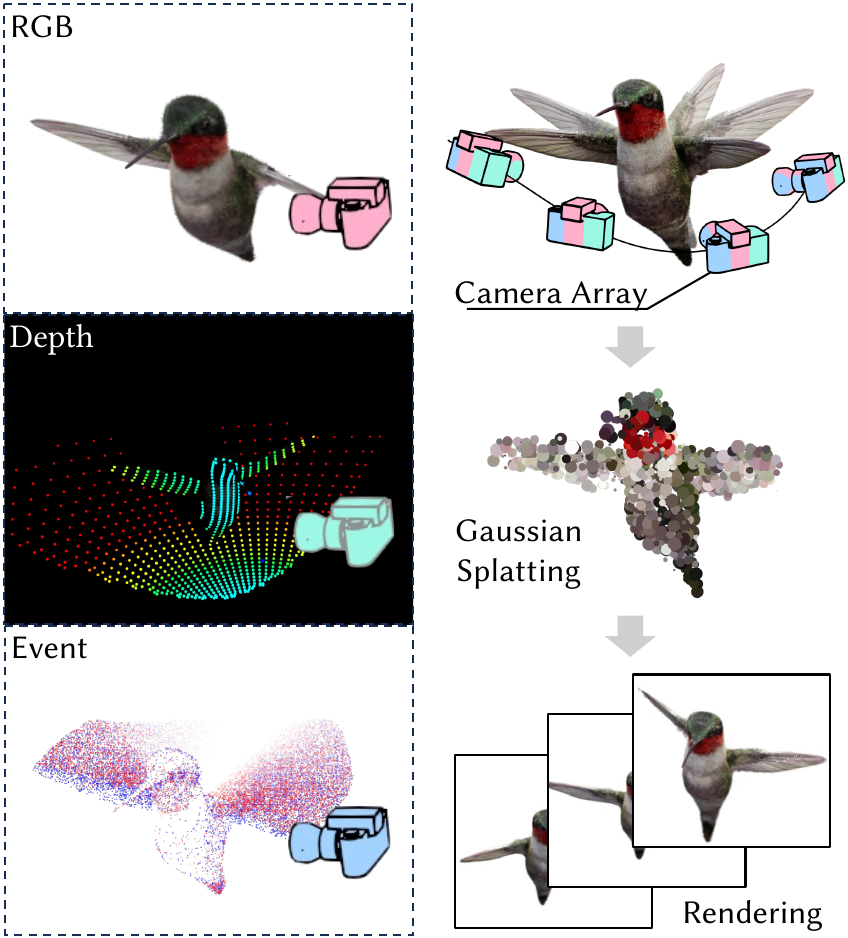}
    \label{fig:overall}
    }
  \subfloat[Applications]{
    \includegraphics[width=0.567\linewidth]{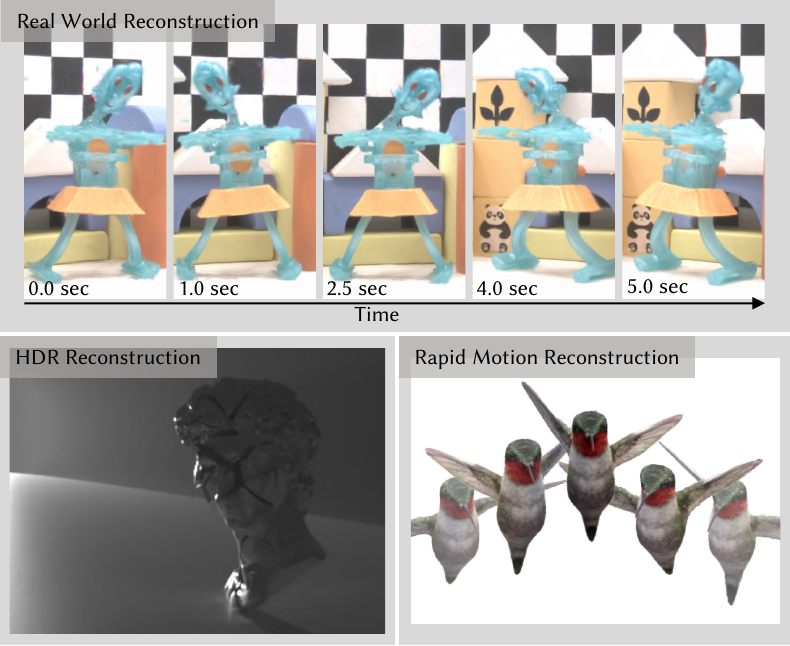}
    \label{fig:real_world_applications}
  }
  \vspace{-1em}
  \caption{\textit{High-speed Imaging with Sensor Fusion Splatting.}
  We propose a method to reconstruct high-speed, deforming 3D scene geometries by integrating data from three complementary imaging modalities---RGB, event, and depth cameras, as shown in (a). 
  Using these inputs, we train a set of deformable Gaussians to dynamically represent the scene. 
  This approach enables detailed, time-resolved reconstructions of the scene in challenging conditions such as low light, high dynamic range or rapid scene motion and from a limited baseline (see (b)), allowing novel view synthesis at any timestamp.
  }
\Description{teaser}
\end{teaserfigure}

\begin{abstract}
    Capturing and reconstructing high-speed dynamic 3D scenes has numerous applications in computer graphics, vision, and interdisciplinary fields such as robotics, aerodynamics, and evolutionary biology. 
However, achieving this using a single imaging modality remains challenging.
For instance, traditional RGB cameras suffer from low frame rates, limited exposure times, and narrow baselines. 

To address this, we propose a novel sensor fusion approach using Gaussian splatting, which combines RGB, depth, and event cameras to capture and reconstruct deforming scenes at high speeds. 
The key insight of our method lies in leveraging the complementary strengths of these imaging modalities:
RGB cameras capture detailed color information, event cameras record rapid scene changes with microsecond resolution, and depth cameras provide 3D scene geometry.
To unify the underlying scene representation across these modalities, 
we represent the scene using deformable 3D Gaussians.
To handle rapid scene movements, we jointly optimize the 3D Gaussian parameters and their temporal deformation fields by integrating data from all three sensor modalities. 
This fusion enables efficient, high-quality imaging of fast and complex scenes, even under challenging conditions such as low light, narrow baselines, or rapid motion. 
Experiments on synthetic and real datasets captured with our prototype sensor fusion setup demonstrate that our method significantly outperforms state-of-the-art techniques, achieving noticeable improvements in both rendering fidelity and structural accuracy.

\end{abstract}

\maketitle

\todolist

\section{Introduction}\feedbackNeeded
\label{intro}

Capturing high-speed 3D phenomena---such as the flapping wings of a hummingbird, explosions, or industrial processes---has wide-ranging applications in science and engineering. 
However, designing imaging systems for these scenarios is challenging due to their extremely high bandwidth demands and limited baseline geometries. 
Ultra-high-speed cameras (e.g., Phantom, Photron, IDT), although can achieve remarkable frame rates, are prohibitively expensive (over \$100k), offer short capture durations (only a few seconds), lack depth information, and generate massive data volumes. 
Currently, no single imaging modality can capture dynamic scenes with high spatial, temporal, and depth resolutions. 
In this work, we address this limitation by integrating three complementary imaging modalities---RGB cameras, event cameras, and depth cameras---which collectively provide spatial, temporal, and depth information.

RGB cameras are inexpensive and provide high 2D spatial resolution along with color information about the scene. However multiple RGB cameras struggle to achieve high-resolution depth information in limited baseline scenarios~\cite{qadri2024aoneus, qu2024z}, due to the missing cone problem. 
Additionally, RGB cameras face challenges in low-light conditions and are constrained by bandwidth limitations, resulting in low frame rates. 

Event cameras, in contrast, offer ultra-high temporal resolutions (down to microseconds) by capturing only pixel-level changes in brightness. They transmit asynchronous sparse data, known as events, which significantly reduces data redundancy and storage demands. Furthermore, their high dynamic range (HDR) enables them to capture scenes with extreme lighting contrasts, which would otherwise appear washed out or darkened with conventional RGB cameras. 
These features make event cameras an excellent complement to RGB cameras for capturing high-speed scenes.
However, neither of these cameras provide depth information, which complicates reconstruction in small baseline scenarios.

Depth sensors are shown to sample data within the missing cone, capturing essential information that RGB and event cameras cannot \cite{qu2024z}. 
Active depth cameras, such as SPADs \cite{gupta2019photon, folden2024foveaspad, gupta2019asynchronous, jungerman20223d, po2022adaptive} or CWToF sensors \cite{keetha2024splatam, friday2024snapshot, he2019recent, shrestha2016computational}, measure the distance between the camera and the scene, providing necessary depth information for accurate reconstruction, especially in small baseline conditions.

In this work, we integrate information from all three complementary sensory modalities to enable reconstruction of high-speed 3D scenes in small baseline scenarios.
As shown in \figureautorefname{~\ref{fig:overall}}, the image representation of each modality differ significantly: RGB cameras provide color images, event cameras provide pixel-level intensity changes at extremely high speed, and depth cameras generate depth maps.
Fusing this sensory information requires a shared scene representation.
To achieve this, we represent the underlying scene using deformable 3D Gaussians that are shared across all imaging modalities.
We render data from RGB, event and depth cameras through a Gaussian splatting framework, and jointly optimize the 3D Gaussian parameters and their temporal deformation model by
minimizing the differences between the rendered splatting outputs and the observed multi-view, multi-sensory data.
This approach enables reconstruction of high-resolution multiview images and depth maps, achieving enhanced spatial, temporal, and angular fidelity. 
\todo{do we show any multiview images? if not we should.}

We rigorously evaluate our approach on diverse synthetic scenes and real-world scenes captured with our sensor fusion prototype (shown in \figureautorefname{~\ref{fig:hardware_setup}}), including challenging scenarios with rapid motion and low-light conditions---scenarios where traditional cameras or single modalities often fail (see \figureautorefname{~\ref{fig:real_world_applications}}). \todo{do we have real evaluations for all three scenarios?}
Our results demonstrate that the proposed multi-sensor fusion approach significantly improves reconstructions, delivering both fidelity and robustness in capturing high-speed, dynamic scenes.

In summary, we make the following contributions:
\begin{enumerate}
    \item We \textit{propose} Sensor Fusion Splatting, a novel deformable 3D Gaussian splatting framework that integrates RGB, event, and depth data streams to enable robust, high-speed 3D scene reconstruction and consistent photorealistic novel view synthesis from multi-sensor inputs.
    
    \item We \textit{develop} a hardware sensor fusion prototype that combines RGB, event, and depth cameras to capture high-speed deformable 3D scenes, supporting our proposed method. \todo{this point seems like a repetition of the above point}

    \item We rigorously evaluate our approach on diverse synthetic and real-world datasets featuring high-speed, deforming 3D scenes, demonstrating significant performance improvements in scene reconstruction.
\end{enumerate}
\noindent
Our code and data will be made public upon acceptance of the manuscript.

\section{Related Works}
\label{related_works}

We propose a multi-sensor fusion of RGB, event, and depth data using shared deformable Gaussian splatting representation. Our technique builds on recent advances in neural rendering, sensor fusion, and high-speed imaging. 
We outline these advancements and how our method builds upon or differs from them. 
For further details, we refer readers to resources on neural rendering \cite{xie2022neural, tewari2022advances, fei20243d}, event camera applications \cite{gallego2020event}, and depth sensing \cite{horaud2016overview, szeliski2022computer}.

\subsection{Neural Rendering}

Neural radiance field (NeRF) introduced by \citet{mildenhall2021nerf} and its extensions \cite{barron2021mip,zhang2020nerf++} represent 3D scenes implicitly using multilayer perceptrons (MLPs) for inverse differentiable rendering via analysis-by-synthesis.
This method has been adapted to dynamic scenes \cite{pumarola2020d, liu2022devrf} by conditioning NeRF on time \cite{li2021neural, du2021neural, xian2021space} or learning time-based transformations \cite{pumarola2021d,park2021nerfies,tretschk2021non,lombardi2019neural,gao2021dynamic},
with additional performance enhancements via incorporating explicit representations \cite{muller2022instant, mobilenerf, plenoxels, kulhanek2023tetra}.
More recent methods enhance efficiency through plane projection \cite{cao2023hexplane,fridovich2023k,shao2023tensor4d}, compact hashing \cite{wang2024masked}, neural point clouds \cite{kappel2024d}, and separated static and dynamic scene components \cite{song2023nerfplayer}.
Despite these improvements, \nerf-based methods demand numerous neural network evaluations per ray, especially in the empty regions, limiting training speed, and real-time rendering capability. While acceleration techniques like instant-ngp \cite{muller2022instant, chen2025neural} mitigate some inefficiencies,  
implicit representations remain slow and lack explicit deformability.

Recently, 3D Gaussian Splatting (3DGS) \cite{kerbl20233d} has demonstrated remarkable performance in computing the scene parameters, achieving real-time rendering with state-of-the-art quality. 
This method represents a 3D scene with a set of Gaussians with differentiable attributes such as position, scale, rotation, density, and radiance.
Since its introduction, 3DGS has been extended in several ways,
with improvements targeting anti-aliasing \cite{mip-splatting}, dynamic scene handling \cite{yang2023gs4d, duan20244d}, and optimized rendering for large scenes \cite{hierarchGS}; also see overview provided by \cite{fei20243d}.
Of particular relevance to this work are Gaussian splatting techniques adapted to dynamic scenes, such as per-frame Gaussian models \cite{luiten2024dynamic}, continuous deformations via neural networks \cite{xie2024gaussian, yang2024deformable}, deformation fields on planes \cite{wu20244d}, and trajectory modeling of Gaussians using polynomial and Fourier series fitting \cite{gao2024gaussianflow}. 
In this work, we leverage 3D Gaussians as a shared scene representation across different sensor modalities and utilize Gaussian splatting techniques for 3D reconstruction of deforming scenes from complementary multisensory data.

\subsection{High-speed Imaging}

Event cameras are sensors that transmit binary events per pixel, indicating an increase or decrease of the observed brightness. 
Their high dynamic range and rapid temporal resolution make them ideal for computational photography \cite{han2020neuromorphic, tulyakov2021time, tulyakov2022time}. 
Research on 2D applications of event cameras, including full-frame reconstruction \cite{kim2008simultaneous, rebecq2016evo, scheerlinck2018continuous} and deblurring \cite{haoyu2020learning, sun2022event}, have already seen extensive progress, while 3D imaging applications have started emerging at rapid pace in recent times \cite{hidalgo2020learning,baudron2020e3d,cui2022dense,uddin2022unsupervised,wang2022evac3d,kim2016real,rebecq2018emvs,hwang2023ev,low2023robust}.
Some early works combined event cameras with lasers or structured light projectors for 3D reconstruction \cite{brandli2014adaptive, matsuda2015mc3d}, while more recent research has explored reconstructing 3D scenes by integrating event cameras with stereo setups \cite{uddin2022unsupervised} or LiDAR \cite{cui2022dense}.
Notable progress has also been made in 3D event-based reconstruction using NeRF and 3DGS to achieve high multi-view coherence in dynamic scenes \cite{rudnev2023eventnerf,klenk2023nerf,hwang2023ev,low2023robust,ma2023deformable,xiong2024event3dgs,guo2024spikegs}.
Yet, these systems often struggle with the limited photometric data captured by event cameras, which provide only derivative information based on motion.
We overcome these challenges by fusing event cameras with RGB and depth sensors, and demonstrate high-speed imaging across diverse and challenging scenarios.

\begin{figure*}[ht!]
  \centering
  \includegraphics[width=\linewidth]{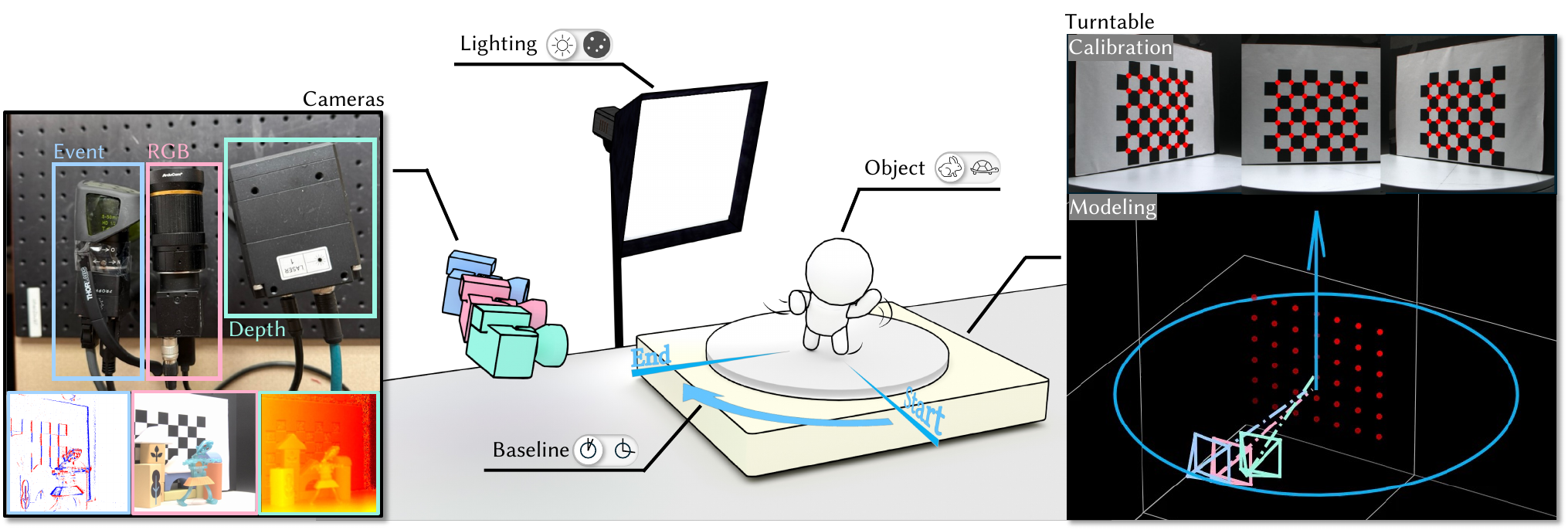}
  \vspace{-2em}
  \caption{\textit{Sensor Fusion Imaging Prototype.} Our sensor fusion system consists of RGB, event and depth cameras providing complementary sensory information. A high-speed deforming object placed on a rotating turntable is imaged by these cameras, as illustrated, and the proposed sensor fusion splatting approach is used for reconstruction. The complementary modality sensors and turntable are calibrated to ensure accurate alignment. \todo{improve time illustration.}}
  \label{fig:hardware_setup}
  \Description{hardware-setup}
\end{figure*}

\subsection{Sensor Fusion}
To address the limitations inherent to using single sensing modality, sensor fusion and multimodal sensing algorithms have been widely explored.
Combining complementary data sources, such as depth and RGB images, has been extensively studied for refining depth estimation, leveraging the ubiquity of RGB-D cameras \cite{henry2012rgb, bleiweiss2009fusing, boominathan2014improving}. 
Radar fusion with cameras has enabled applications in object detection \cite{nabati2021centerfusion}, imaging in cluttered environments \cite{grover2001low}, and autonomous navigation. 
Lidar and optical time-of-flight sensors have similarly been used to image through scattering media \cite{bijelic2020seeing}, densify lidar scans \cite{kim2009multi}, and improve depth estimation \cite{lindell2018single, nishimura2020disambiguating, attal2021torf}.
To overcome the limitations of optical cameras, researchers have also fused optical sensors with acoustic sensors such as microphone arrays \cite{chakravarthula2023seeing,lindell2019acoustic} and sonars \cite{Ferreira2016,qadri2024aoneus,williams2004simultaneous,Raaj2016,babaee20153}.
Sensor fusion using event-based cameras is relatively a nascent field, with most approaches targeting applications of RGB and event camera fusion \cite{feng2020deep,liang2019multi,gehrig2021combining} for overcoming RGB-only camera limitations. 
Building on these concepts, we develop a sensor fusion approach integrating RGB, event and depth cameras to achieve high-speed capture of rapid real-world motion.

\section{Sensor Fusion for High-Speed 3D Imaging}\feedbackNeeded
\label{method}

The data provided by our multisensor fusion setup includes 
a set of RGB image frames $\Ical=\{\Ibm\}$, events $\Ecal=\{\ebm\}$, and depth maps $\Dcal=\{\Dbm\}$.
Given these,
our goal is to reconstruct the underlying high-speed, dynamic 3D scene. 
In \subsectionautorefname{~\ref{preliminary}}, we provide a brief overview of event camera data modeling and 3D Gaussian splatting.
In \subsectionautorefname{~\ref{subsec:event-splatting}}, we present our sensor fusion splatting framework, which enables joint optimization of 3D Gaussians and their temporal deformation fields for reconstructing high-speed, dynamic 3D scenes from multi-sensor data.
The details of our hardware imaging prototype are discussed in \sectionautorefname{~\ref{real_world_exp}}.

\subsection{Preliminaries}
\label{preliminary}

\noindent
\textbf{Modeling Event Camera}. 
An event camera operates with a grid of independent pixels, each responding to changes in the log photocurrent. 
In particular, each pixel asynchronously generates an event $\ebm_k = (\xbm_k, t_k, p_k)$ at time $t_k$ when the difference of logarithmic brightness $L$ at the pixel $\xbm_k = (x_k, y_k)^T$ reaches a predefined threshold $C$:
\begin{equation}
    L(\xbm_k, t_k) - L(\xbm_k, t_k - \Delta t_k) = p_k C,
\end{equation}
where 
$p\in\left\{- 1, + 1\right\}$ is the sign (or polarity) of the brightness change, and $\Delta t_k$ is the time since the last event at the pixel $\xbm_k$. 
The result is a sparse sequence of events which are asynchronously triggered by radiance changes.
For simplicity, we refer to an event at position $\xbm$ and time $t$ as $\ebm_{\xbm,t}$, with its corresponding polarity denoted as $p_{\xbm,t}$.

\noindent
\\
\textbf{3D Gaussian Splatting}. \gs~represents a 3D scene explicitly as a set of anisotropic 3D Gaussians, with 
each Gaussian defined by its mean position $\mubm\in\R^3$ and covariance matrix $\Sigmabm\in\R^{3\times3}$.
To enable differentiable optimization, the covariance matrix is decomposed as $\Sigmabm=\Rbm\Sbm\Sbm^\top\Rbm^\top$, where $\Sbm\in\R^3_+$ is the scaling matrix and $\Rbm\in SO(3)$ is the rotation matrix. 
We simplify these matrices using the 
quaternion $\rbm\in\R^4$ for rotations and the scaling factor $\sbm\in\R^3$. 
For image rendering, the 3D Gaussians are projected (splatted) onto a 2D camera plane, and the mean position $\mubm$ and covariance matrix $\Sigmabm$ are transformed using
the following equations:
\begin{equation}
\mubm^{\prime}=\Pbm \Wbm \mubm,
\end{equation}
\begin{equation}
\Sigmabm^{\prime}=\Jbm\Wbm \Sigmabm \Wbm^\top \Jbm^\top,
\end{equation}
where $\mubm^{\prime}$ and $\Sigmabm^\prime$ are the projected 2D mean position and covariance matrix, and $\Pbm$, $\Wbm$, and $\Jbm$ represent the projective transformation, viewing transformation, and the Jacobian of the affine approximation of $\Pbm$, respectively. 
After projection, we apply alpha blending to calculate the final pixel value $C(\xbm)$ at pixel $\xbm$ by blending the $N$ ordered Gaussian along the ray shot from the pixel:
\begin{equation}
\label{alpha_blending}
C(\xbm)=\sum_{i \in N} \cbm_i \sigma_i \prod_{j=1}^{i-1}\left(1-\sigma_j\right),
\end{equation}
where $\cbm_i$ denotes the view-dependent radiance,
$\sigma_i$ denotes the density of each Gaussian, and $\prod(\cdot)$ denotes the cumulative visibility.
Finally, given a set of target images, Gaussian splatting frames 3D reconstruction as an inverse problem, and iteratively refines the parameters of 3D Gaussians $G(\mubm, \rbm, \sbm, \cbm, \sigma)$ to minimize the difference between projected splats and actual image data, resulting in high-fidelity scene reconstructions.

\subsection{Sensor Fusion Splatting}
\label{subsec:event-splatting}

As illustrated in \figureautorefname{~\ref{fig:overall}}, our method leverages RGB for color, event cameras for high-speed motion, and depth sensors for spatial structure, to
enable robust, high-quality 3D reconstructions of fast-moving, complex scenes. 
Three tailored loss functions ensure precise and consistent optimization across all sensor data.

\vspace{1mm}
\noindent
\textbf{Modeling Deformable Scenes}.
Instead of training separate sets of 3D Gaussians for each time-dependent RGB view and interpolating between them, we capture continuous motion and reconstruct 3D scenes with spatio-temporal consistency.
By decoupling motion from geometry, we map 3D Gaussians into a canonical space, creating time-independent representations guided by geometric priors.
These priors link 3D Gaussian positions to spatio-temporal changes, supervised using multi-sensor data.
To efficiently encode scene dynamics, we employ a grid encoder \cite{cao2023hexplane, wu20244d}, which maps spatial and temporal features onto 2D grids, and thereby accelerating rendering.

Specifically, we sample the encoded features from a given time $t$ and the 3D Gaussian parameters $\mubm$ as inputs, and decode the offset values that deform the 3D Gaussians $\Delta G$ using a multilayer perceptron (MLP) network. 
This deformation is given as:
\begin{equation}
   \Delta G(\mubm,\sbm,\rbm) = M_{\gbm, \thetabm}(\mubm, \gamma(t)),
   \label{eq:deformation-field}
\end{equation}
where $M_{\gbm, \thetabm}$ denotes the MLP producing the deformation field, $\gbm$ denotes the grid features, ${\thetabm}$ represents the MLP parameters and 
where 
$\gamma(t) = (\sin(2^k \pi t), \cos(2^k \pi t))$ denotes positional encoding.
Novel images are then generated by splatting the deformed 3D Gaussians updated with parameters obtained from the deformation field.

\vspace{1mm}
\noindent 
\textbf{Event Supervision}. 
We splat the 3D Gaussians based on the events provided by the event camera and the deformable scene modeling discussed above, to reconstruct high-speed dynamic scenes.
Specifically, integrating the event information, which provides dense log intensity changes over time, in optimizing the scene parameters complement the temporal sparsity of traditional RGB or depth cameras.
To optimize the Gaussians using temporal events, we start by randomly selecting two timestamps, $t_s$ and $t_e$ (with $t_s<t_e$), from a given set of events. 
We model the ground truth log-intensity difference $\Delta\Lbm^*\in\R^{H\times W}$, between $t_s$ and $t_e$ as:
\begin{equation}
\Delta\Lbm_{t_s\veryshortarrow t_e}(\xbm)=\begin{cases}
\int_{t_s}^{t_e} \eta p_{\xbm,t}\diff t & \text{if } \{\ebm_{\xbm,t}\mid t_s<t<t_e\}\ne \emptyset \\
0 & \text{otherwise}
\end{cases},
\end{equation}
where $\eta$ is the contrast threshold, and $p_{\xbm,t}$ represent the event polarity.
For supervising and optimizing temporally consistent Gaussian deformations, we also predict the log-intensity difference between splatted images $\{\widehat{\Ibm}\}$ at the selected times $t_s$ and $t_e$:
\begin{equation}
\Delta \widehat{\Lbm}_{t_s\veryshortarrow t_e} = \log{\widehat{\Ibm}_{t_e}}-\log{\widehat{\Ibm}_{t_s}}.
\end{equation}

A key consideration to note while simulating event data is that the time window $t_e-t_s$ significantly impacts reconstruction quality: shorter windows inhibit propagation of broader lighting changes, while longer windows compromise fine details due to event neutralization -- consecutive events of opposite polarity canceling each other -- inevitably caused during the signal integration time.
This was also observed in some previous studies simulating event data \cite{rudnev2023eventnerf,low2023robust,xiong2024event3dgs}.
To address this constraint, we adopt two strategies:
(1) we randomly sample the window length as $t_e-t_s\sim\Ucal[l_{min},l_{max}]$ to balance temporal resolution with event density ($l_{min}$ and $l_{max}$ being the minimum and maximum window lengths) , and 
(2) to prevent misinterpretation of neutralized events during integration, we apply a mask to exclude these canceled events from loss calculation. This approach ensures they are not mistaken for zero events, avoiding unnecessary penalization for actual intensity variations.
Therefore, the optimization objective for splatting based on event signals is given as:
\begin{equation}
\Lcal_{event} = |\Delta\Lbm_{t_s\veryshortarrow t_e} - \Delta \widehat{\Lbm}_{t_s\veryshortarrow t_e}|_2\odot\mbm,
\label{eq:event-splat}
\end{equation}
where $\mbm$ denotes the event mask.

While directly supervising with raw event data has been widely used for \gs~and \nerf~training \cite{klenk2023nerf, rudnev2023eventnerf, low2023robust, xiong2024event3dgs}, raw events in real-world, multi-sensor, and dynamic scenarios can become corrupted by noise and misaligned due to calibration errors. 
Empirically, we observe that directly training with raw events introduce wavering artifacts and jumpy background, ultimately degrading the rendering performance.

To address these issues, we additionally supervise our training using reconstructed frames from the E2VID method \cite{rebecq2019high}. 
We observe that naively minimizing MSE or L1 loss between rendered \gs~frames and reconstructed E2VID frames yields temporal inconsistencies and fog-like artifacts in the rendered frames due to inherent calibration errors and misalignments.\todo{may need to show these artifacts in sup and ref here}
To mitigate these artifacts and improve temporal consistency, we employ the Learned Perceptual Image Patch Similarity (LPIPS) metric \cite{zhang2018unreasonable} as a perceptual loss $\Lcal{lpips}$:

\begin{equation} 
\Lcal_{lpips} = \text{LPIPS}(\Ibm_t, \widehat{\Ibm}_t),
\end{equation}
where, at a time $t$, $\Ibm_t$ is reconstructed from events via E2VID, and $\widehat{\Ibm}_t)$ is rendered at the same time $t$ using \equationautorefname{\ref{alpha_blending}}. 
By aligning the \gs~generated frames with E2VID reconstructions, we reduce visual artifacts in renderings and enhance temporal stability.
Since E2VID produces only grayscale images, we integrate RGB camera data into the pipeline to provide the necessary color information.

\vspace{1mm}
\noindent 
\textbf{RGB Supervision}.
Given a ground truth image $\Ibm_t\in\R^{H\times W\times 3}$ at time $t$, an image $\widehat{\Ibm}_t$ is rendered via 3DGS using \equationautorefname{~\ref{alpha_blending}}. The objective function for training 3D Gaussians $G$ and MLP (parameterized by ${\thetabm}$) with given multi-view RGB images is given by

\begin{equation}
\Lcal_{rgb} = |\Ibm_t-\widehat{\Ibm}_t|.
\end{equation}
\noindent
This RGB image supervision provides necessary color information.

\vspace{1mm}
\noindent
\textbf{Depth Supervision}. 
To ensure structural consistency throughout the spatio-temporal changes in the scene, we incorporate supervision using depth maps from a depth sensor. To this end, we render the depthmap $\widehat{\Dbm}_t$ using \equationautorefname{~\ref{alpha_blending}}, but replace $\cbm$ with the z distance of the gaussians in camera space.
The depth-based loss function is defined as:

\begin{equation}
\Lcal_{depth}=|\Dbm_t-\widehat{\Dbm}_t|.
\end{equation}

where $\Dbm_t\in\R^{H\times W}_{+}$ represents the ground truth depth map at time $t$. 
Incorporating depth splatting encourages accurate depth alignment during reconstruction of fast-moving scenes, reinforcing structural consistency across frames.

\vspace{1mm}
\noindent
\textbf{Overall Loss Function}. 
In addition to the aforementioned objectives, we apply a second-order temporal smoothness regularization \cite{cao2023hexplane} to the deformation grid features $\gbm$ from \equationautorefname{~\ref{eq:deformation-field}}.
This regularization prevents artifacts and enhances temporal smoothness of reconstructed high-speed scenes, and is computed as:
\begin{equation}
    \Lcal_{\gbm} = \|\nabla^2_{t}\gbm\|^2
\end{equation}
The final objective function for our approach is given by:

\begin{equation}
\Lcal=\lambda_1 \Lcal_{rgb}+\lambda_2 \Lcal_{event} + \lambda_3 \Lcal_{lpips} + \lambda_4 \Lcal_{depth} + \lambda_5 \Lcal_{\gbm},
\label{eq:final-objective}
\end{equation}
where the $\lambda$s are the weight balancing factors.

\section{Results}
\label{experiments}

\begin{table*}[!ht]
\caption{\textit{Quantitative Evaluations on Synthetic Data.} We evaluate our method on simulated scenes with rapid motion and compare it against existing relevant approaches. 
Our method consistently outperforms competing techniques across all evaluated scenes, highlighting the benefits of sensor fusion splatting. The best performance for each metric is highlighted in \textbf{bold}.}
\centering
\small %
\begin{tabularx}{\linewidth}{l>{\centering\arraybackslash}X>{\centering\arraybackslash}X>{\centering\arraybackslash}X|>{\centering\arraybackslash}X>{\centering\arraybackslash}X>{\centering\arraybackslash}X|>{\centering\arraybackslash}X>{\centering\arraybackslash}X>{\centering\arraybackslash}X|>{\centering\arraybackslash}X>{\centering\arraybackslash}X>{\centering\arraybackslash}X}
\hline
\multirow{2}{*}{Methods} & \multicolumn{3}{c}{Hummingbird} & \multicolumn{3}{c}{Rotating Balls} & \multicolumn{3}{c}{Sculpture}  & \multicolumn{3}{c}{Lego}\\ \cline{2-13} 
                         & \makebox{\hspace{-3.5pt}PSNR}   & \makebox{\hspace{-3.5pt}LPIPS}   & \makebox{\hspace{-4pt}DRMS}  & \makebox{\hspace{-3.5pt}PSNR}   & \makebox{\hspace{-3.5pt}LPIPS}   & \makebox{\hspace{-4pt}DRMS} & \makebox{\hspace{-3.5pt}PSNR}   & \makebox{\hspace{-3.5pt}LPIPS}   & \makebox{\hspace{-4pt}DRMS} & \makebox{\hspace{-3.5pt}PSNR}   & \makebox{\hspace{-3.5pt}LPIPS}   & \makebox{\hspace{-4pt}DRMS}\\ \hline
Robust e-NeRF\cite{low2023robust}      & 14.01      & 0.435      & 9.0921     & 24.11  & 0.190  & 2.160    &19.98 &0.400 &0.599 & 8.97  & 0.484  & 4.954\\
\makecell[l]{Deformable \gs\cite{yang2024deformable}}      & 30.50      & 0.041      & 2.014    & 36.92  & 0.021  & 2.133  & 32.26  & 0.079  & 1.776 & 30.12 & 0.033 & 0.206 \\
4DGS\cite{wu20244d}                     & 25.37  & 0.076  & 2.150  & 32.68  & 0.025  & 1.301 & 32.06  & 0.074  & 2.152 & 30.10 & 0.041 & 0.281 \\
\ours      & \textbf{36.45}      &   \textbf{0.030}    & \textbf{1.615}     & \textbf{39.82}  & \textbf{0.012}  & \textbf{0.051}  & \textbf{34.82}  & \textbf{0.071}  & \textbf{0.346}  & \textbf{32.49}  & \textbf{0.025}  & \textbf{0.038} \\ \hline

\end{tabularx}
\label{tab:synthesis_data}
\end{table*}

This section evaluates our method on synthetic and real-world data. 
Experimental setup and analysis for simulated data are detailed in \subsectionautorefname{~\ref{syn_exp}}, while hardware prototype details and results for real-world dynamic scenes are in \subsectionautorefname{~\ref{real_world_exp}}. 
In \subsectionautorefname{~\ref{analysis}}, we analyze the impact of scene conditions---such as viewing angles, lighting, motion speed, and depth map quality---on performance and discuss our optimization strategies 
across dynamic scenarios.

\vspace{1mm}
\noindent
\textbf{Optimization Details}
Our method, implemented in PyTorch, is tested on an NVIDIA V100 GPU. 
We train our sensor-fusion \gs~approach using the Adam optimizer \cite{diederik2014adam}, fine-tuning parameters as outlined by \citet{kerbl20233d}. 
For the deformation field in \equationautorefname{~\ref{eq:deformation-field}}, 
we model deformations of $\mubm$, $\sbm$, and $\rbm$, scaling the time dimension of the deformation grid based on the degree of scene motion.
Event supervision (\equationautorefname{~\ref{eq:event-splat}}) uses temporal window length of $l_{min}=1$ ms across all scenes, with $l_{max}$ adjusted for illumination---smaller for bright scenes and larger for dim ones.
To optimize the final objective in \equationautorefname{~\ref{eq:final-objective}}, we initialize with a static vanilla \gs~model trained for 3000 steps, followed by training the full sensor-fusion model for 30,000 steps.

\subsection{Synthetic Evaluations}
\label{syn_exp}

\noindent
\textbf{Simulation and Synthetic Data.}
To evaluate our approach under diverse dynamic scene conditions, particularly those challenging to capture in real-world environments, we developed a simulator combining Blender \cite{blender} and ESIM \cite{rebecq2018esim}. 
Our synthetic dataset includes three custom scenes and one scene from \citet{ma2023deformable},
rendered at 1000fps and a resolution of $400\times 400$ pixels using Blender. Depth information was directly extracted from Blender, while events were simulated using ESIM.

We categorized synthetic scenes based on three characteristics:
baseline (large, small), lighting(bright, dark), and object speed(slow, fast). 
The scenes are:

\begin{itemize}
    \item \textit{\textbf{Hummingbird}}. 
    A hummingbird flaps its wings at very high speeds while the camera moves in a small circle around it. The baseline is small, lighting is bright, and object motion is fast.
    \item \textit{\textbf{Rotating Balls}}. In a dark setting, a red ball orbits a green ball, both self-rotating.
    The baseline is small, lighting is dark, and object motion is slow.
    \item \textit{\textbf{Sculpture}}. A sculpture disintegrates rapidly while the camera orbits around it.
    The baseline is large, lighting is bright, and object motion is fast
    \item \textit{\textbf{Lego}}. From \citet{ma2023deformable}, this scene features a Lego bulldozer with the bucket performing two sweeping motions, while the camera orbits around it.
    The baseline is large, lighting is bright, and object motion is slow.
\end{itemize}

\begin{figure*}[!ht]
    \centering
    \includegraphics[width=0.97\linewidth]{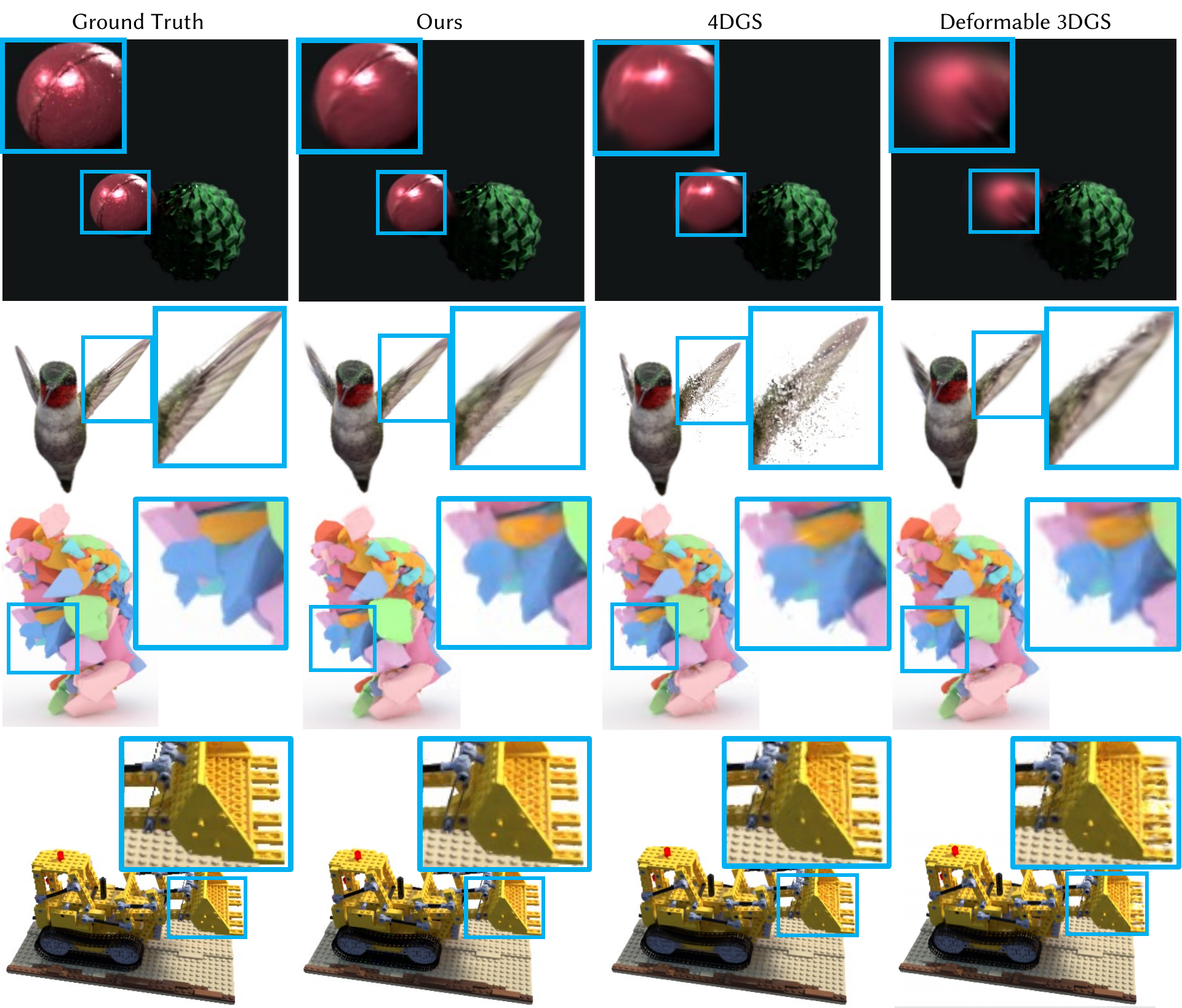}
    \caption{\textit{Synthetic Evaluations}.
    We visualize four dynamic scenes synthesized using our method and two competing baselines (Deformable \gs\cite{yang2024deformable} and 4DGS\cite{wu20244d}).
    Our method consistently outperforms the baseline methods capturing the finer appearance details. 
    }
    \label{fig:general-results}
    \Description{general-result\todo{}}
\end{figure*}

\vspace{1mm}
\noindent
\textbf{Experiments and Assessment}.
We evaluate our experimental results using several metrics, including per-pixel accuracy (PSNR), perceptual quality (LPIPS)\cite{zhang2018unreasonable}, and depth root mean square (DRMS).
We compare our method to Deformable 3DGS\cite{yang2024deformable}, 4DGS\cite{wu20244d}, and Robust e-NeRF\cite{low2023robust}. For Robust e-NeRF\cite{low2023robust}, since we only have monochrome event data, we calculate the metrics in grayscale.
Quantitative results for our synthetic dataset are summarized in \tableautorefname{~\ref{tab:synthesis_data}}, and qualitative results are presented in \figureautorefname{~\ref{fig:general-results}}. As indicated in \tableautorefname{~\ref{tab:synthesis_data}}, our method consistently outperforms other methods across all quality metrics. Notably, in the visual results in \figureautorefname{~\ref{fig:general-results}}, our approach captures fine details in dynamic scenes, such as the accurate reconstruction of surfaces like the specular ball in column 1, and rapidly moving hummingbird's wings in column 2, where other methods falter.

\begin{figure}[!ht]
    \centering
    \includegraphics[width=\columnwidth]{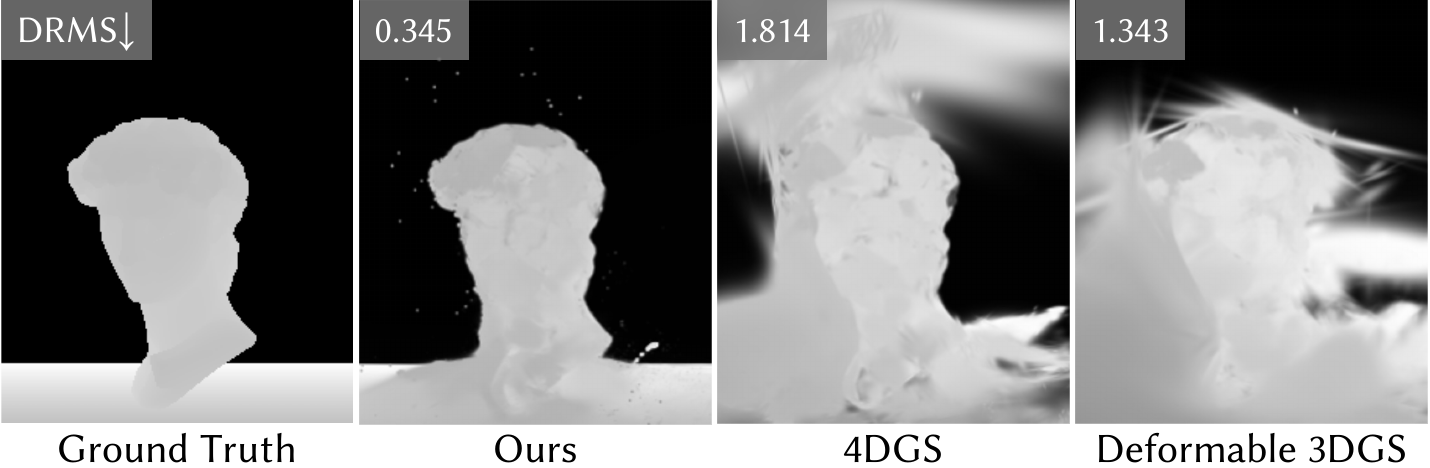}
    \vspace{-2em}
    \caption{\textit{Comparison of Depth Map Reconstructions}. 
    We present a rendered depth map from an unseen camera viewpoint.
    Our method achieves the closest approximation to the ground truth depth compared to existing approaches.
    }
    \label{fig:depth}
    \Description{depth}
\end{figure}

\begin{figure}[!t]
    \centering
    \includegraphics[width=\columnwidth]{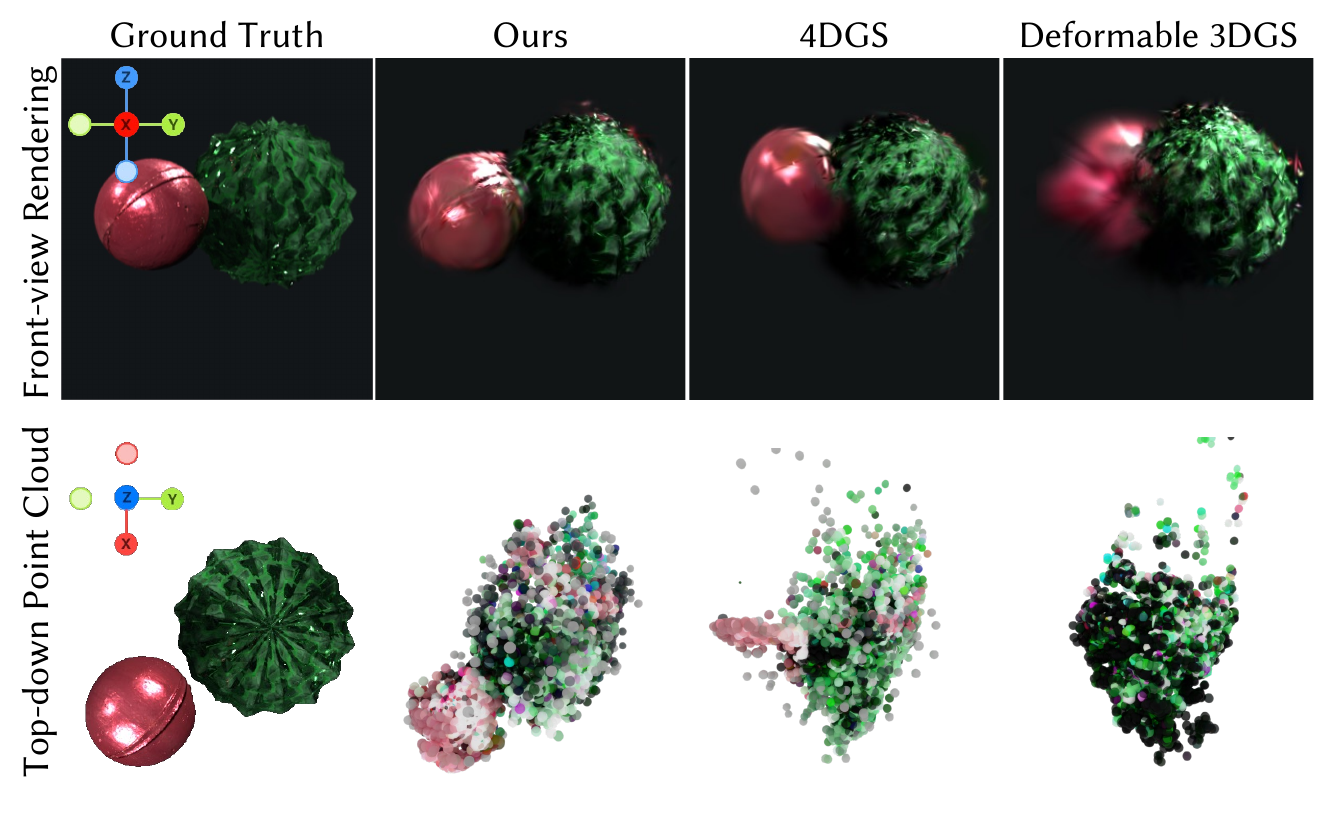}
    \vspace{-2em}
    \caption{\textit{Evaluating Structural Accuracy}. 
    We present images rendered from a viewpoint outside the training camera positions and a top-down view of the extracted point cloud. 
    Our method accurately positions the red ball and captures its 3D structure, unlike other methods that either misplace it, represent it incompletely, or merge it with the green ball.
    }
    \label{fig:extrapolate}
    \Description{depth}
    \vspace{-4mm}
\end{figure}

\noindent
\textbf{Structural Accuracy}. Depth supervision enhances the accuracy of the 3D structure in our method. In \figureautorefname{~\ref{fig:depth}}, we present a depth map rendered from a viewpoint interpolated between training cameras. This supervision allows our method to effectively learn the 3D structure, distinguishing the object from the background more clearly than other methods. In \figureautorefname{~\ref{fig:extrapolate}}, we show images rendered from a viewpoint outside the training camera positions, along with a top-down view of the point cloud extracted from the Gaussians. Our method accurately positions the small red ball, whereas other methods either misplace it or represent it incompletely. The point cloud reveals that our method successfully captures the 3D structure of the red ball, unlike other methods that depict it as either a flat circle or merge it with the green ball.

\subsection{Real-World Evaluations}
\label{real_world_exp}

\noindent
\textbf{Hardware Prototype}. To validate our method, we built a hardware prototype consisting of three sensors: a FLIR Blackfly S machine vision camera for high-resolution RGB data, a a Prophesee EVK4 event-based camera for high-speed motion information, and LUCID Helios2+ depth camera for capturing spatial structure. These cameras are mounted around a speed-adjustable turntable, allowing precise control over scene rotation. A programmable LED light is positioned above the turntable to provide adjustable illumination. \figureautorefname{~\ref{fig:hardware_setup}} illustrates our hardware setup and the integration of the cameras with the turntable.

\vspace{1mm}
\noindent
\textbf{Real-World Data}. We capture three distinct scenes using the aforementioned hardware prototype. We capture the scenes for 5 seconds under 30 frame per second. We crop the RGB frames to size of 900$\times$900, the event data to size of 600$\times$600, and the depth data to size of 380$\times$380. The captured scenes are:
\begin{itemize} 
    \item \textit{\textbf{Dancing Toy}}: In a bright setting, a windup toy dances rapidly with its head waving in a dizzying speed. We set the turntable to rotate 72 degrees. The baseline is large, the lighting is bright, and the object speed is fast. 
    \item \textit{\textbf{Dancing Toy (Dark)}}: We repeat the \textit{\textbf{Dancing Toy}} scene, but with a dim illumination.
    \item \textit{\textbf{Newton's Cradle}}: A Newton's cradle bounces in a bright setting. We set the turntable to rotate 37.5 degrees. The baseline is small, the lighting is bright, and the object speed is fast. 
\end{itemize}

\begin{figure*}[!ht]
    \centering
    \includegraphics[width=\linewidth]{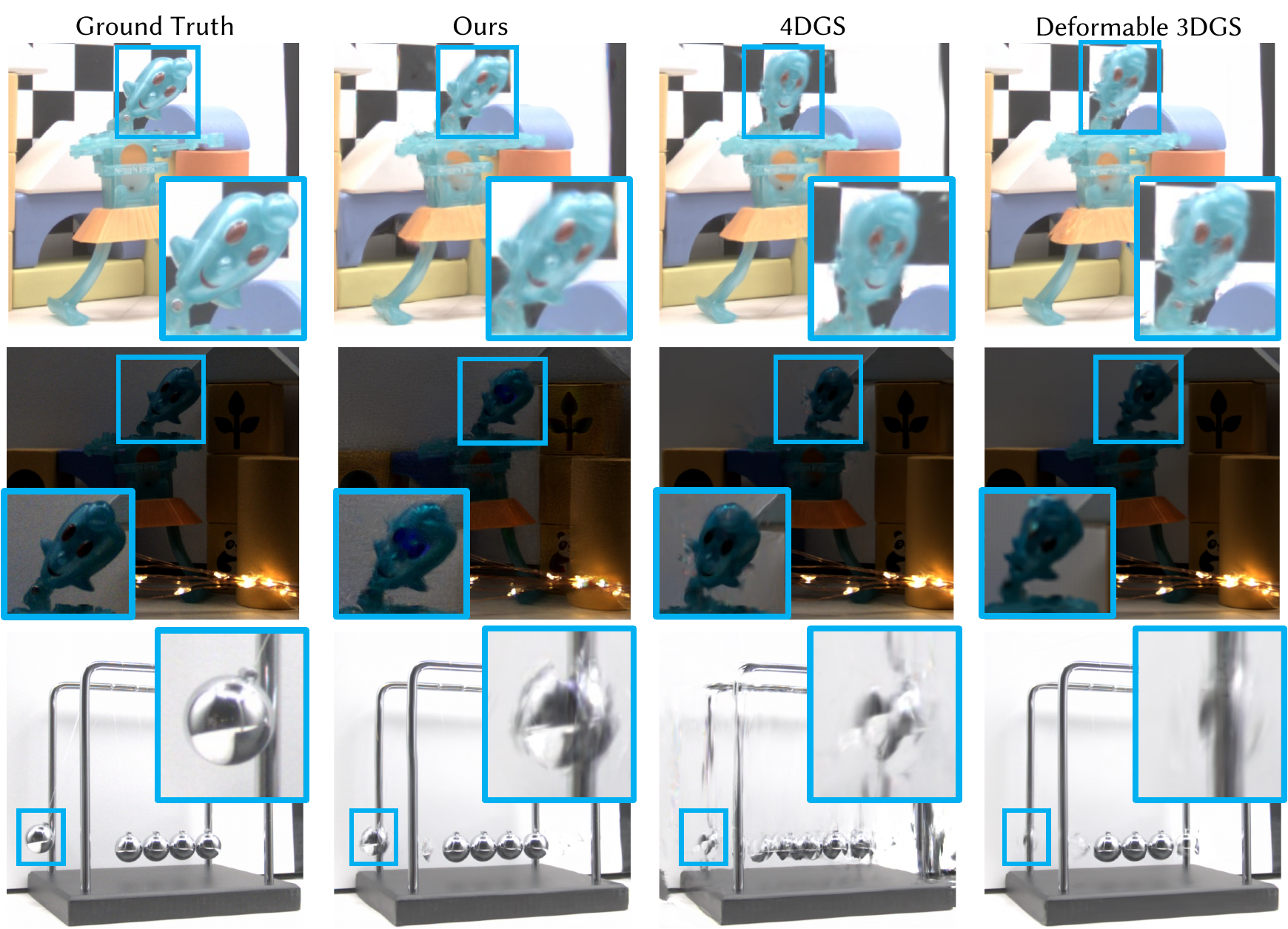}
    \caption{\textit{Real-world Evaluations}. We present three dynamic scenes reconstructed with our method and two competing baselines (Deformable \gs\cite{yang2024deformable} and 4DGS\cite{wu20244d}). 
    Our method not only achieves superior visual quality but also maintains temporal consistency, whereas baseline methods misinterpret object motion during interpolation.
    \todo{capture the dancing toy scene with dimmer lighting. replace row two.}\todo{decrease font size}}
    \label{fig:real_world_visualization}
    \Description{real_world_visualization}
\end{figure*}

\noindent
\textbf{Experiments}. We visualize the reconstructed results in \figureautorefname{~\ref{fig:real_world_visualization}}. In the \textit{\textbf{Dancing Toy}} scene, our method achieves smoother motion interpolation, correctly maintaining the toy's head position, in contrast to other methods that position it incorrectly. For the \textit{\textbf{Dancing Toy (Dark)}} scene, our method consistently reconstructs smooth motion even in low-light conditions, unlike other methods that introduce artifacts around the neck or blur the face. However, we observe a color jitter effect around the eyes, which is a common phenomenon in monochromatic event-based reconstruction, as reported in \cite{xiong2024event3dgs,ma2023deformable}. For the \textit{\textbf{Newton's Cradle}} scene, the reconstruction of the specular surfaces of the balls is a significant challenge for \gs, as discussed in \cite{fan2024spectromotion, yang2024spec}. Despite these difficulties, our method successfully reconstructs the balls and ensures smooth interpolation between frames, whereas the other methods either introduce artifacts or completely omit the balls.\todo{add the dark scene discussion if we have it}

\subsection{Analysis}
\label{analysis}

\begin{figure*}[]
    \centering
    \subfloat[Varying Baselines]{
    \includegraphics[width=0.52\linewidth]{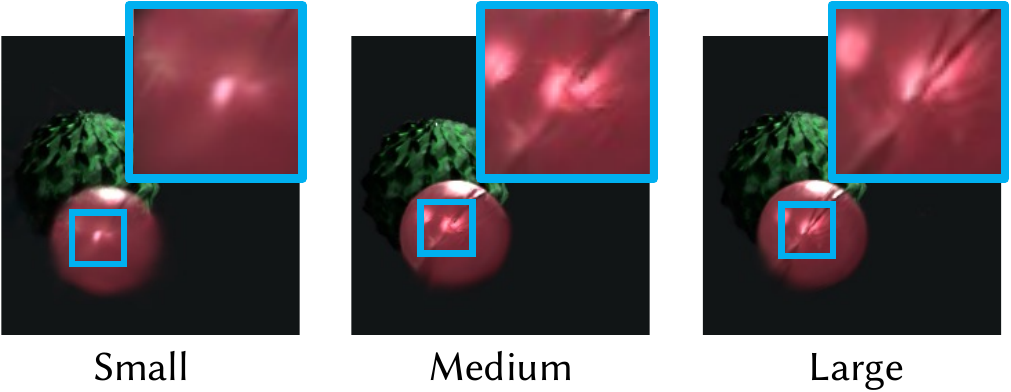}
    \label{fig:baseline}
    }
    \subfloat[Varying Number of Training Frames]{
    \includegraphics[width=0.465\linewidth]{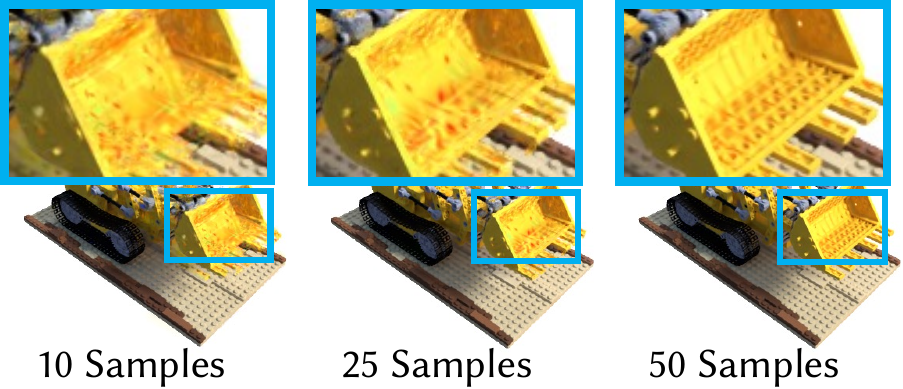}
    \label{fig:num_frames}
    }
    \caption{\textit{Evaluation under Varying Baselines and Training Frames}. 
    We demonstrate the visual performance of our method under varying baselines and number of training frames. 
    Our method achieves robust reconstruction quality even with small baselines and maintains high performance with sparse training samples.}
    \label{fig:baseline_numframes}
    \Description{effect}
\end{figure*}

\begin{figure*}[]
    \centering
    \subfloat[Varying Object Speed]{
    \includegraphics[width=0.5\linewidth]{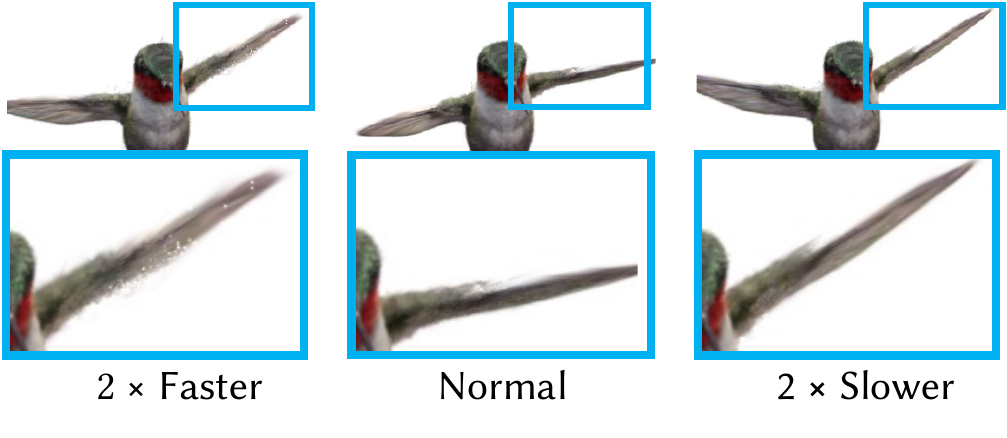}\vspace{-0.73em}
    \label{fig:object_speed}
    }
    \subfloat[Varying Lighting]{
    \includegraphics[width=0.47\linewidth]{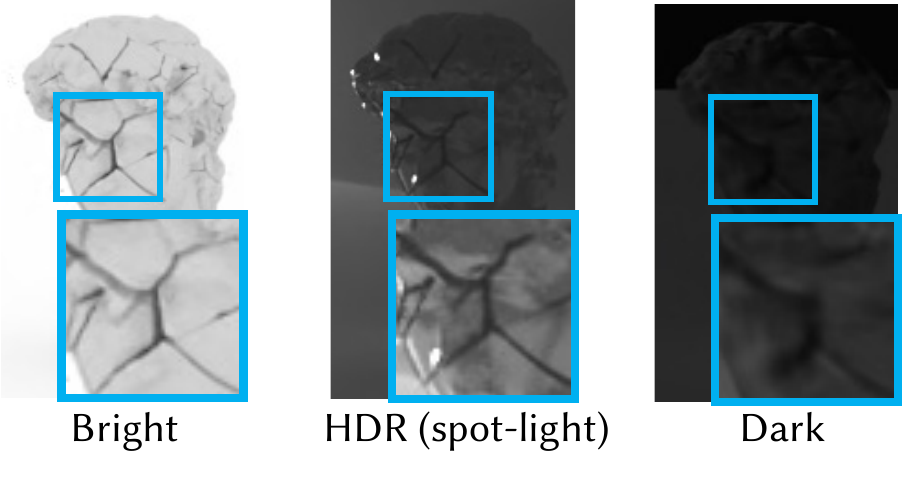}
    \label{fig:lighting}
    }
    \caption{\textit{Evaluation under Varying Object Speeds and Lighting}. 
    We demonstrate the visual performance of our method under varying object speed and lighting conditions. 
    Our method achieves robust reconstruction quality even in scenarios with very fast motion and challenging lighting conditions, including bright, dark and high-dynamic range illumination.}
    \label{fig:object_speed_lighting}
    \Description{effect}
\end{figure*}

Imaging quality is constrained by factors such as limited camera baselines, low lighting, or slow capture speeds. Reconstructing scenes under such extreme conditions is critical for many applications. Here, we demonstrate the robustness of our method under these limiting conditions. In \tableautorefname{~\ref{tab:conditions}}, we compare our method with 4DGS \cite{yang2023gs4d} under different conditions. In \figureautorefname{~\ref{fig:baseline_numframes}} and \figureautorefname{~\ref{fig:object_speed_lighting}}, we visualize the comparison across different conditions of our method.

\begin{table}[]
\setlength{\tabcolsep}{1pt}
\caption{\textit{Quantitative Evaluations under Varying Conditions}. Our method constantly outperforms existing baselines (Deformable 3DGS \cite{yang2024deformable} and 4DGS \cite{wu20244d}) with significant improvements in visual reconstruction quality.}
\label{tab:conditions}
\centering
\small %
\begin{tabularx}{\linewidth}{lXXXXXX}
\hline
\textbf{Methods} & \multicolumn{6}{c}{\textbf{Conditions}}  \\\hline
& \multicolumn{1}{c|}{\hspace{2mm}PSNR} & DRMS&  \multicolumn{1}{c|}{\hspace{3mm}PSNR}&DRMS&  \multicolumn{1}{c|}{\hspace{3mm}PSNR}& DRMS \\\hline
\multicolumn{7}{c}{\hspace{27mm}\textit{Camera Baselines}}\\\hdashline
& \multicolumn{2}{c}{large} & \multicolumn{2}{c}{medium} & \multicolumn{2}{c}{small} \\\hdashline

\makecell[l]{Deformable 3DGS} & \multicolumn{2}{c}{\hspace{-0.7mm}36.92~|~2.133} & \multicolumn{2}{c}{35.94~|~2.422} & \multicolumn{2}{c}{35.78~|~2.498}\\
4DGS  & \multicolumn{1}{r|}{32.68}& \multicolumn{1}{l}{1.301} &   \multicolumn{2}{c}{32.74~|~1.258} &  \multicolumn{2}{c}{31.06~|~1.760} \\
Ours & \multicolumn{1}{r|}{\textbf{39.82}}&\multicolumn{1}{l}{\textbf{0.051}} & \multicolumn{2}{c}{\textbf{38.20}~|~\textbf{0.066}} & \multicolumn{2}{c}{\textbf{36.49}~|~\textbf{0.097}}\\\hline

\multicolumn{7}{c}{\hspace{27mm}\textit{Number of Training Samples}}\\\hdashline
& \multicolumn{2}{c}{50} & \multicolumn{2}{c}{25} & \multicolumn{2}{c}{10} \\\hdashline

\makecell[l]{Deformable 3DGS} & \multicolumn{2}{c}{\hspace{-0.1mm}30.12~|~0.206} & \multicolumn{2}{c}{24.14~|~1.380} & \multicolumn{2}{c}{20.81~|~1.686}\\
4DGS  & \multicolumn{2}{c}{30.10~|~0.281} & \multicolumn{2}{c}{25.32~|~2.446} & \multicolumn{2}{c}{20.37~|~2.365} \\
Ours & \multicolumn{2}{c}{\textbf{32.49}~|~\textbf{0.038}} & \multicolumn{2}{c}{\textbf{30.48}~|~\textbf{0.047}} & \multicolumn{2}{c}{\textbf{25.36}~|~\textbf{0.115}} \\\hline

\multicolumn{7}{c}{\hspace{27mm}\textit{Object Speed}}\\\hdashline
& \multicolumn{2}{c}{2$\times$faster} & \multicolumn{2}{c}{normal} & \multicolumn{2}{c}{2$\times$slower} \\\hdashline

\makecell[l]{Deformable 3DGS} & \multicolumn{2}{c}{\hspace{-0.1mm}24.82~|~3.000} & \multicolumn{2}{c}{30.50~|~2.014} & \multicolumn{2}{c}{30.44~|~2.621}\\
4DGS  & \multicolumn{2}{c}{23.38~|~3.277} & \multicolumn{2}{c}{25.37~|~2.150} & \multicolumn{2}{c}{26.61~|~2.371} \\
Ours & \multicolumn{2}{c}{\textbf{34.45}~|~\textbf{1.432}} & \multicolumn{2}{c}{\textbf{36.45}~|~\textbf{1.615}} & \multicolumn{2}{c}{\textbf{36.69}~|~\textbf{1.674}} \\\hline

\multicolumn{7}{c}{\hspace{27mm}\textit{Lighting}}\\\hdashline
& \multicolumn{2}{c}{bright} & \multicolumn{2}{c}{spot-light} & \multicolumn{2}{c}{dark} \\\hdashline
\makecell[l]{Deformable 3DGS} & \multicolumn{2}{c}{\hspace{-0.1mm}36.27~|~1.762} & \multicolumn{2}{c}{31.38~|~0.747} & \multicolumn{2}{c}{38.42~|~1.385}\\
4DGS  & \multicolumn{2}{c}{35.50~|~2.191} & \multicolumn{2}{c}{31.66~|~0.813} & \multicolumn{2}{c}{40.29~|~0.972} \\
Ours & \multicolumn{2}{c}{\textbf{41.37}~|~\textbf{0.259}} & \multicolumn{2}{c}{\textbf{41.40}~|~\textbf{0.381}} & \multicolumn{2}{c}{\textbf{43.04}~|~\textbf{0.423}} \\\hline

\end{tabularx}
\end{table}

\noindent
\textbf{Camera Baseline}. 
We evaluate the impact of different camera baselines on scene reconstruction quality using the \textit{\textbf{Rotating Balls}} scene, captured with \textit{large}, \textit{medium}, and \textit{small} baselines.
As shown in \tableautorefname{~\ref{tab:conditions}} and \figureautorefname{~\ref{fig:object_speed}}, our method achieves robust reconstruction quality even with the smallest baseline.

\vspace{1mm}
\noindent
\textbf{Number of Training Samples}. 
We investigate the performance of our method with varying number of training samples on the \textit{\textbf{Lego}} scene.
The results are presented in \tableautorefname{~\ref{tab:conditions}} and \figureautorefname{~\ref{fig:num_frames}}. 
Unlike other methods, which exhibit a significant decline in structural accuracy with fewer frames, our sensor fusion method remains robust even with sparse training samples.

\vspace{1mm}
\noindent
\textbf{Object Speed}. 
We evaluate reconstruction quality for objects moving at different speeds using the \textit{\textbf{Hummingbird}} scene, where the hummingbird flaps its wings at %
$2\times$ faster and $2\times$ slower speeds. 
As shown in \tableautorefname{\ref{tab:conditions}} and \figureautorefname{\ref{fig:object_speed}}, our method outperforms others in image quality, even at faster speeds. 
We observe that the structural accuracy modestly decreases at slower wing speeds due to increased overlap of the wings behind the body, making it more challenging  
for the Gaussians to accurately predict the 3D structure.

\vspace{1mm}
\noindent
\textbf{Lighting}. 
RGB camera performance often deteriorates under extreme lighting conditions, such as low or high dynamic range lighting, making robust reconstruction essential for many applications. 
In the \textit{\textbf{Sculpture}} scene, we evaluate our method under three lighting scenarios: \textit{Bright} (sufficient lighting), \textit{Dark} (low lighting), and \textit{Spotlight} (one side illuminated by a spotlight, the other side in shadow). 
The sculpture's color is unified to marble white across all scenes, with increased metallic appearance in the \textit{Spotlight} scene to enhance HDR effects.
As shown in \tableautorefname{~\ref{tab:conditions}} and \figureautorefname{~\ref{fig:lighting}}, our method demonstrates robust reconstruction across these challenging lighting conditions---see the reconstruction of surface cracks in all three cases. 
Note that in the \textit{Dark} setting, the PSNR for all methods is higher due to much of the sculpture appearing as a uniform black color, reducing detail variations. 
However, as seen in \figureautorefname{~\ref{fig:lighting}}, the visual quality in the \textit{Dark} setting is not necessarily better than under other lighting conditions.

\section{Conclusion}
\label{conclusion}

We introduced a novel sensor fusion method for imaging high-speed dynamic 3D scenes. 
By integrating RGB, depth and event cameras, and a novel sensor fusion splatting method, we are able to efficiently image and reconstruct rapid, complex scenes 
while significantly reducing storage and bandwidth requirements compared to traditional high-speed imaging techniques and sensors.
Our approach demonstrates superior performance under complex and challenging conditions, including low light, low contrast, and fast motion, as validated through both synthetic and real-world experiments.
Leveraging a point-based scene representation, the proposed method supports efficient imaging, modeling, and rendering of dynamic scenes, making it highly suitable for a variety of high-speed imaging and post-production applications.

\bibliographystyle{templates/acmart/acmart}
\bibliography{bibliography}

\clearpage

\end{document}